\documentclass[USenglish,twocolumn]{article}

\usepackage[utf8]{inputenc}
\usepackage[big]{dgruyter}
\usepackage{graphicx}

\begin{document}


\author*[1]{Tosin P. Adewumi} 

\author[2]{Foteini Liwicki}
\author[3]{Marcus Liwicki}

\affil[1]{SRT Department, EISLAB, Luleå University of Technology, 97187, Sweden, E-mail: tosin.adewumi@ltu.se}

\affil[2]{SRT Department, EISLAB, Luleå University of Technology, 97187, Sweden, E-mail: foteini.liwicki@ltu.se}

\affil[3]{SRT Department, EISLAB, Luleå University of Technology, 97187, Sweden, E-mail: marcus.liwicki@ltu.se}
\title{\huge Word2Vec: Optimal hyper-parameters and their impact on NLP downstream tasks}

  \runningtitle{Article title}


 \begin{abstract}
{Word2Vec is a prominent model for natural language processing (NLP) tasks.
Similar inspiration is found in distributed embeddings for new state-of-the-art (SotA) deep neural networks.
However, wrong combination of hyper-parameters can produce poor quality vectors.
The objective of this work is to empirically show optimal combination of hyper-parameters exists and evaluate various combinations.
We compare them with the released, pre-trained original word2vec model.
Both intrinsic and extrinsic (downstream) evaluations, including named entity recognition (NER) and sentiment analysis (SA) were carried out.
The downstream tasks reveal that the best model is usually task-specific, high analogy scores don't necessarily correlate positively with F1 scores and the same applies to the focus on data alone.
Increasing vector dimension size after a point leads to poor quality or performance.
If ethical considerations to save time, energy and the environment are made, then reasonably smaller corpora may do just as well or even better in some cases.
Besides, using a small corpus, we obtain better WordSim scores, corresponding Spearman correlation and better downstream performances (with significance tests) compared to the original model, trained on a 100 billion-word corpus.
}
\end{abstract}


  \startpage{1}

  \journalyear{2016}
  \journalvolume{1}
  \journalissue{11}

\maketitle
\section{Introduction}
There have been many implementations of the word2vec model in either of the two architectures it provides: continuous skipgram and continuous bag-of-words (CBoW) \cite{mikolov2013efficient}.
Similar distributed models of word or subword embeddings (or vector representations) find usage in SotA, deep neural networks like bidirectional encoder representations from transformers (BERT) and its successors \cite{devlin2018bert,liu2019roberta,raffel2019exploring}.
BERT generates contextual representations of words after been trained for extended periods on large corpora, unsupervised, using the attention mechanisms \cite{vaswani2017attention}.
Unsupervised learning provides feature representations using large unlabelled corpora \cite{langkvist2014review}.

It has been observed that various hyper-parameter combinations have been used in different research involving word2vec, after its release, with the possibility of many of them being sub-optimal \cite{dhingra2017comparative,naili2017comparative, wang2018comparison}.
Therefore, the authors seek to address the research question: what is the optimal combination of word2vec hyper-parameters for intrinsic and extrinsic NLP purposes, specifically NER and SA?
There are astronomically high numbers of combinations of hyper-parameters possible for neural networks, even with just a few layers \cite{levy2015improving}.
Hence, the scope of our extensive, empirical work over three English corpora is on dimension size, training epochs, window size and vocabulary size for the training algorithms (hierarchical softmax and negative sampling) of both skipgram and CBoW.

The objective of this work is to determine the optimal combinations of word2vec hyper-parameters for intrinsic evaluation (semantic and syntactic analogies) and a few extrinsic evaluation tasks \cite{zhang2019biowordvec,wang2019evaluating}.
It is not our objective in this work to set new SotA results.
Some main contributions of this research are the empirical establishment of optimal combinations of word2vec hyper-parameters for NLP tasks, discovering the behaviour of quality of vectors vis-a-vis increasing dimensions and the confirmation of embeddings performance being task-specific for the downstream.
The rest of this paper is organised as follows: related work, materials and methods used, experimental that describes experiments performed, results and discussion that present final results, and conclusion.

\section{Related Work}
Breaking away from the non-distributed (high-dimensional, sparse) representations of words, typical of traditional bag-of-words or one-hot-encoding \cite{turian2010word}, \cite{mikolov2013efficient} created word2vec.
Word2Vec consists of two shallow neural network architectures: continuous skipgram and CBoW.
It uses distributed (low-dimensional, dense) representations of words that group similar words. 
This new model traded the complexity of deep neural network architectures, by other researchers, for more efficient training over large corpora.
Its architectures have two training algorithms: negative sampling and hierarchical softmax \cite{mikolov2013distributed}.
The released model was trained on Google news dataset of 100 billion words.
Implementations of the model have been undertaken by researchers in the programming languages Python and C++, though the original was done in C \cite{rehurek_lrec}.
The Python implementations are slower to train, being an interpreted langauge \cite{adewumi2018inner,adewumi2019inner}.

Continuous skipgram 
predicts (by maximizing classification of) words before and after the center word, for a given range.
Since distant words are less connected to a center word in a sentence, less weight is assigned to such distant words in training.
CBoW, on the other hand, uses words from the history and future in a sequence, with the objective of correctly classifying the target word in the middle.
It works by projecting all history or future words within a chosen window into the same position, averaging their vectors.
Hence, the order of words in the history or future does not influence the averaged vector.
This is similar to the traditional bag-of-words.
A log-linear classifier is used in both architectures \cite{mikolov2013efficient}.
In further work, they extended the model to be able to do phrase representations and subsample frequent words \cite{mikolov2013distributed}.
Earlier models like latent dirichlet allocation (LDA) and latent semantic analysis (LSA) exist and effectively achieve low dimensional vectors by matrix factorization \cite{deerwester1990indexing,levy2015improving}.

It's been shown that word vectors are beneficial for NLP tasks \cite{turian2010word}, such as SA and NER.
Besides, \cite{mikolov2013efficient} showed with vector space algebra that relationships among words can be evaluated, expressing the quality of vectors produced from the model.
The famous, semantic example: \textit{vector("King") - vector("Man") + vector("Woman") $\approx$ vector("Queen")} can be verified using cosine distance.
Syntactic relationship examples include plural verbs and past tense, among others.
WordSimilarity-353 (WordSim) test set is another analysis tool for word vectors \cite{finkelstein2002placing}.
Unlike Google analogy score, which is based on vector space algebra, WordSim is based on human expert-assigned semantic similarity on two sets of English word pairs.
Both tools measure embedding quality, with a scaled score of 1 being the highest (very much similar or exact, in Google analogy case).

Like word embeddings, subword representations have proven to be helpful when dealing with out-of-vocabulary (OOV) words and \cite{thomason2020jointly} used such embeddings to guide the parsing of OOV words in their work on meaning representation for robots.
Despite their success, word embeddings display biases (as one of their shortcomings) seen in the data they are trained on \cite{bolukbasi2016man}.
Intrinsic tests, in the form of word similarity or analogy tests, reveal meaningful relations among words in embeddings, given the relationship among words in context \cite{mikolov2013efficient,pennington2014glove}.
However, it is inappropriate to assume such intrinsic tests are sufficient in themselves, just as it is inappropriate to assume one particular downstream test is sufficient to generalise the performance of embeddings on all NLP tasks \cite{gatt2018survey,adewumi2020corpora,adewumi2020challenge}.

\cite{mikolov2013efficient} tried various hyper-parameters with both architectures of their model, ranging from 50 to 1,000 dimensions, 30,000 to 3,000,000 vocabulary sizes, 1 to 3 epochs, among others.
In our work, we extended research to 3,000 dimensions and epochs of 5 and 10.
Different observations were noticed from the many trials.
They observed diminishing returns after a certain point, despite additional dimensions or larger, unstructured training data.
However, quality increased when both dimensions and data size were increased together.
Although they pointed out that choice of optimal hyper-parameter configurations depends on the NLP problem at hand, they identified the most important factors as architecture, dimension size, subsampling rate, and the window size.
In addition, it has been observed that larger datasets improve the quality of word vectors and, potentially, performance on downstream tasks \cite{adewumi2019conversational,mikolov2013efficient}
.

\section{Materials and methods}
\subsection{Datasets}
The corpora used for word embeddings are the 2019 English Wiki News Abstract by \cite{wikidumpAbstract} of about 15MB, 2019 English Simple Wiki (SW) Articles by \cite{wikidumpSimple} of about 711MB and the Billion Word (BW) of 3.9GB by \cite{41880}.
The corpus used for sentiment analysis is the internet movie database (IMDb) of movie reviews by \cite{maas2011learning} while that for NER is the Groningen Meaning Bank (GMB) by \cite{bos2017groningen}, containing 47,959 sentence samples.
The IMDb dataset used has a total of 25,000 sentences with half being positive sentiments and the other half being negative sentiments.
The GMB dataset has 17 labels, with 9 main labels and 2 context tags.
Google (semantic and syntactic) analogy test set by \cite{mikolov2013efficient} and WordSimilarity-353 (with Spearman correlation) by \cite{finkelstein2002placing} were chosen for intrinsic evaluations.

\subsection{Embeddings}
The hyper-parameters tuned in a grid search for the embeddings are given in table \ref{tab1}.
The models were generated in a shared cluster running Ubuntu 16 with 32 CPUs of 32x Intel Xeon 4110 at 2.1GHz.
Gensim \cite{rehurek_lrec} Python library implementation of word2vec was used.
This is because of its relative stability, popular support and to minimize the time required in writing and testing a new implementation in Python from scratch.
Our models are available for confirmation and source codes are available on github.\footnote{https://github.com/tosingithub/sdesk}

\subsection{Downstream Architectures}
The downstream experiments were run on a Tesla GPU on a shared DGX cluster running Ubuntu 18.
Pytorch deep learning framework was used.

A long short term memory network (LSTM) was trained on the GMB dataset for NER.
A BiLSTM network was trained on the IMDb dataset for SA.
The BiLSTM includes an additional hidden linear layer before the output layer.
Hyper-parameter details of the two networks for the downstream tasks are given in table \ref{tab2}.
The metrics for extrinsic evaluation include F1, precision, recall and accuracy scores (in the case of SA).

\begin{table}[t]
\centering
\caption{Upstream hyper-parameter choices}
\begin{tabular}{c|c}
\textbf{Hyper-parameter} & \textbf{Values} \\
\hline
Dimension size &  300, 1200, 1800, 2400, 3000 \\
\hline
Window size (w) &  4, 8 \\
\hline
Architecture & Skipgram (s1), CBoW (s0) \\
\hline
Algorithm & H. Softmax (h1), N. Sampling (h0)\\
\hline
Epochs & 5, 10\\
\hline
\end{tabular}
\label{tab1}
\end{table}

\begin{table}[htbp!]
\centering
\caption{Downstream network hyper-parameters}
\begin{tabular}{c|c|c|c|c}
\textbf{Archi} & \textbf{Epochs} & \textbf{Hidden Dim} & \textbf{LR} &
\textbf{Loss} \\
\hline
LSTM &  40 & 128 & 0.01 & Cross Entropy \\
\hline
BiLSTM &  20 & 128 * 2 & 0.0001 & BCELoss \\
\hline
\end{tabular}
\label{tab2}
\end{table}

\begin{figure}[htbp]
   \begin{minipage}{0.5\textwidth}
     \centering
     \includegraphics[width=.9\linewidth]{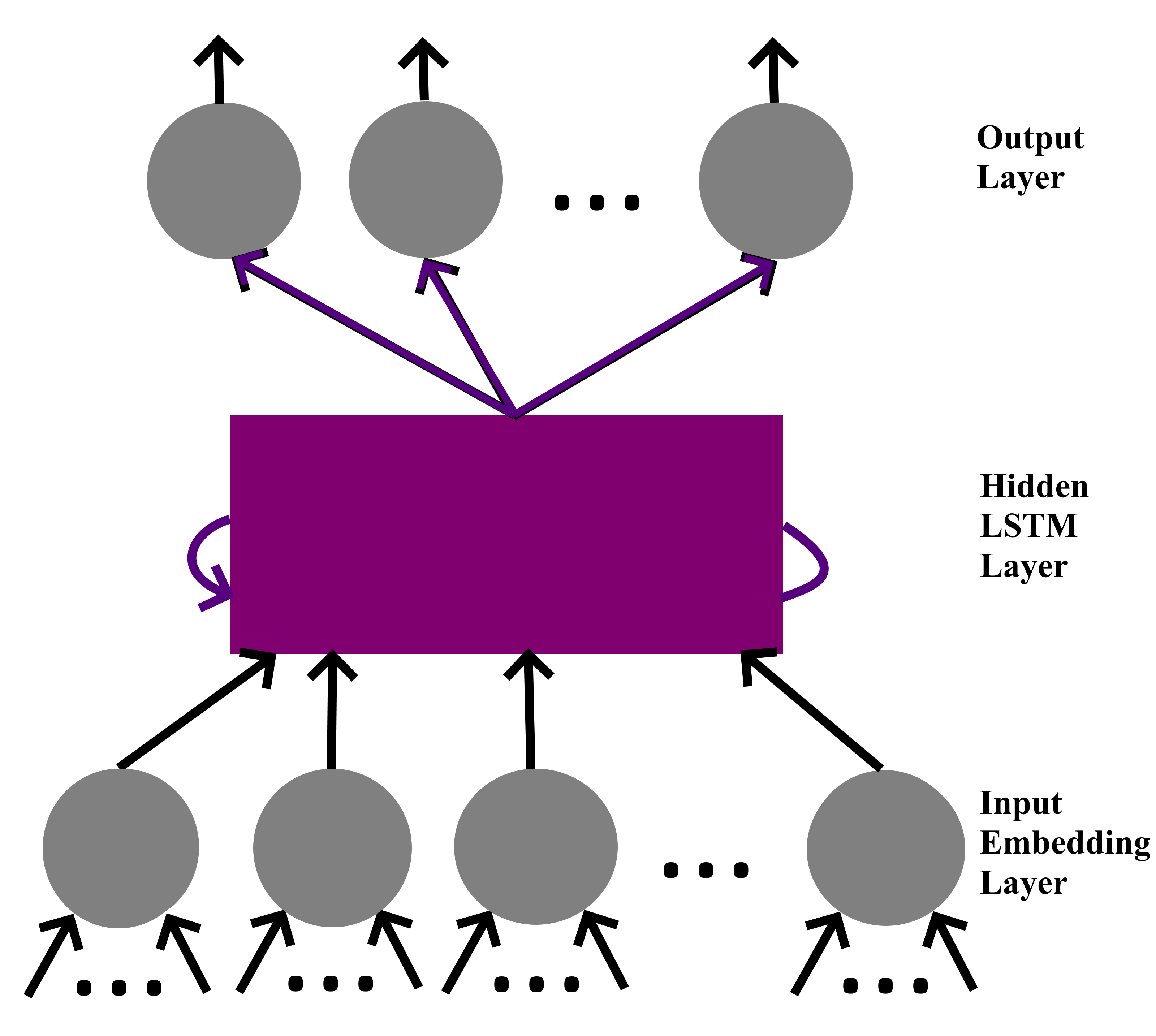}
     \caption{Network architecture for NER}\label{Fig:arch1}
   \end{minipage}\hfill
   \begin{minipage}{0.5\textwidth}
     \centering
     \includegraphics[width=.9\linewidth]{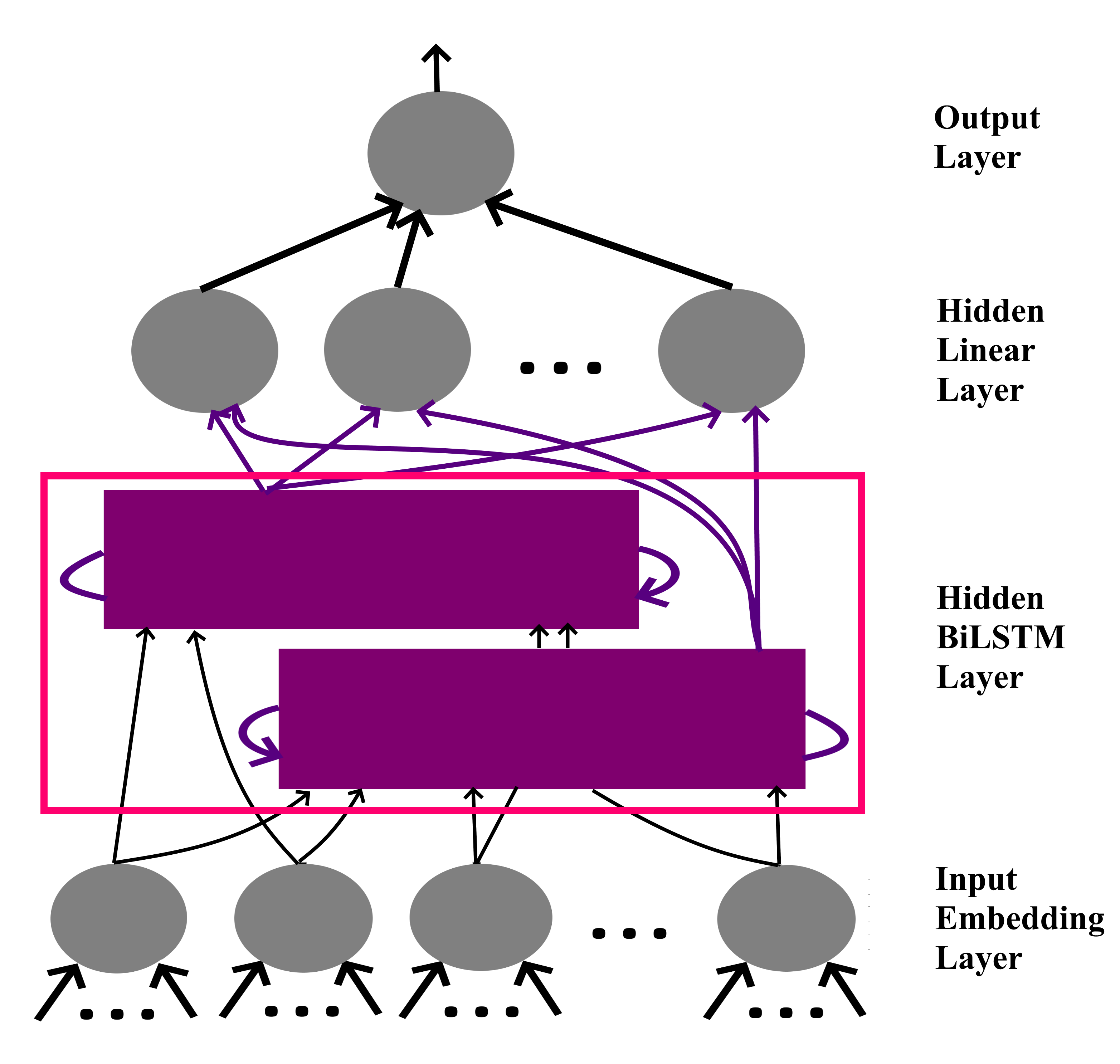}
     \caption{Network architecture for SA}\label{Fig:arch2}
   \end{minipage}
\end{figure}

\section{Experimental}
To form the vocabulary for the embeddings, words occurring less than 5 times in the corpora were dropped, stop words removed using the natural language toolkit (NLTK) \cite{loper2002nltk} and additional data pre-processing carried out.
Table \ref{tab1} describes most hyper-parameters explored for each dataset and notations used.
In all, 80 runs (of about 160 minutes) were conducted for the 15MB Wiki Abstract dataset with 80 serialized models totaling 15.136GB while 80 runs (for over 320 hours) were conducted for the 711MB SW dataset, with 80 serialized models totaling over 145GB.
Experiments for all combinations for 300 dimensions were conducted on the 3.9GB training set of the BW corpus and additional runs for other dimensions for the window size 8 + skipgram + hierarchical softmax combination to verify the trend of quality of word vectors as dimensions are increased.

Preferably, more than one training instance would have been run per combination for a model and an average taken, however, the long hours involved made this prohibitive.
Despite this, we randomly ran a few combinations more than once and confirmed the difference in intrinsic scores were negligible.

For both downstream tasks, the default Pytorch embedding was tested before being replaced by the original (100B) pre-trained embedding and ours.
In each case, the dataset was shuffled before training and split in the ratio 70:15:15 for training, dev and test sets.
Batch size of 64 was used and Adam as optimizer.
For each task, experiments for each embedding was conducted four times and an average value calculated.

\section{Results and Discussion}
\begin{figure}[htbp]
    \centering
    \includegraphics[width=1\linewidth]{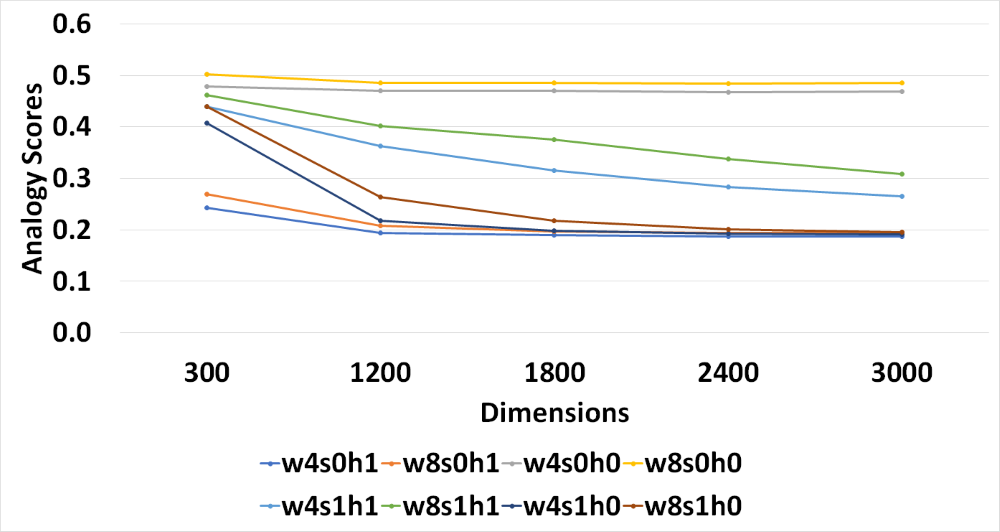}
    \caption{Simple Wiki: Analogy Scores for 10 Epochs \footnotesize{(color needed)}}
    \label{fig:epoch5_wiki1}
\end{figure}

\begin{figure}[htbp]
    \centering
    \includegraphics[width=1\linewidth]{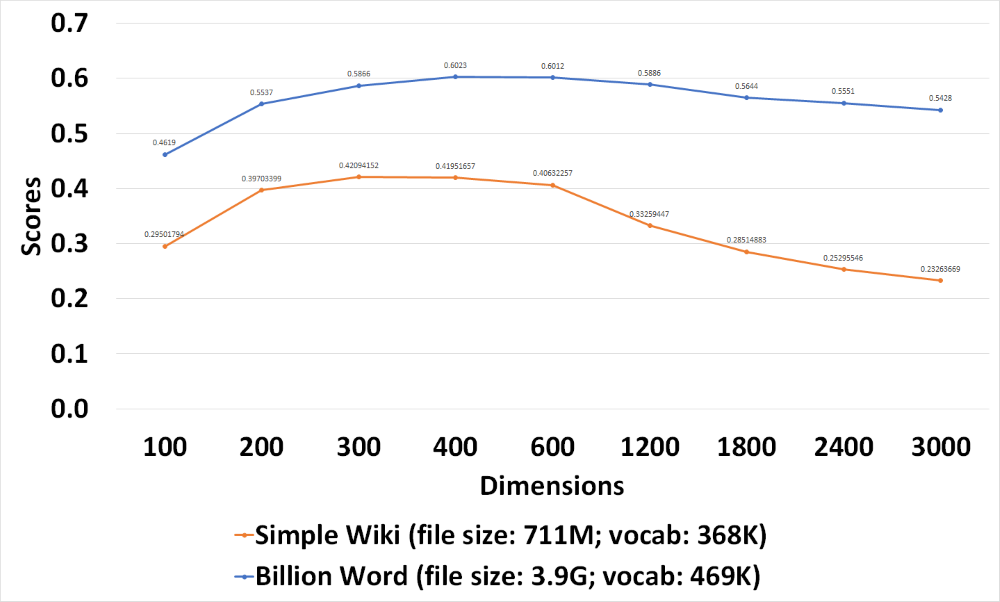}
    \caption{Analogy Scores for w4s1h1 of SW for 5 Epochs \& w8s1h1 of BW for 10 epochs \footnotesize{(not drawn to scale from 400) (color needed)}}
    \label{fig:scores_corpora}
\end{figure}

\begin{figure}[htbp]
    \centering
    \includegraphics[width=1\linewidth]{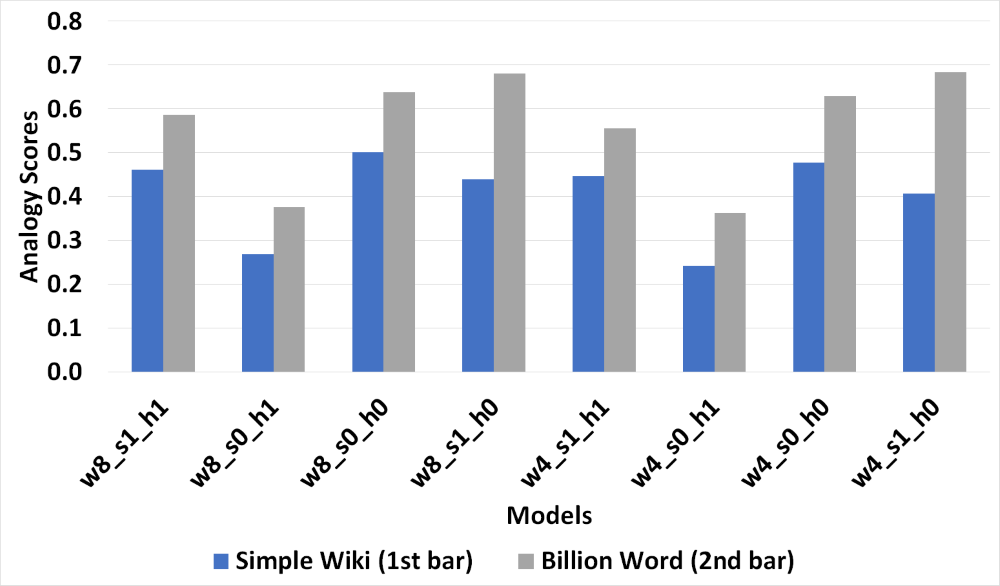}
    \caption{Comparison of 300 dimension models for 10 epochs for SW \& BW corpora}
    \label{fig:corpora_300}
\end{figure}

\begin{figure}[htbp]
    \centering
    \includegraphics[width=1\linewidth]{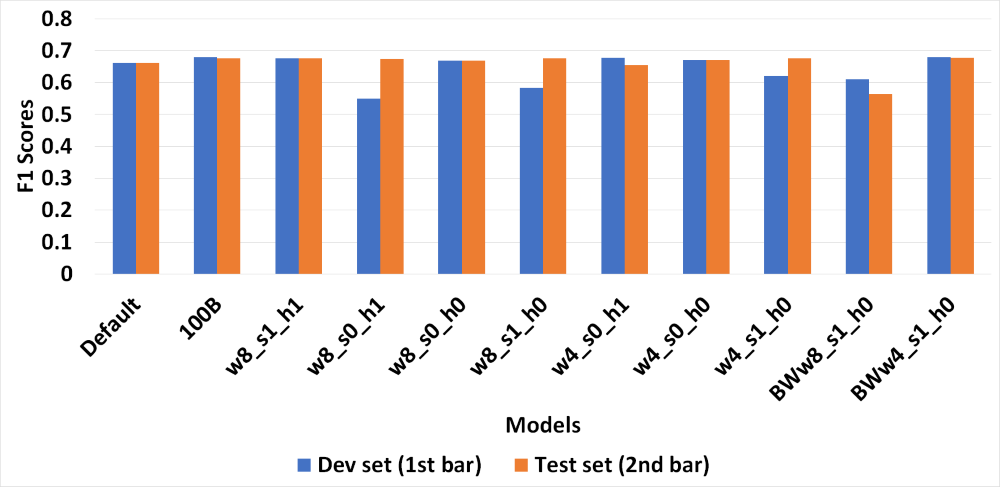}
    \caption{Named Entity Recognition (NER) Mean F1 Scores on GMB Dataset}
    \label{fig:ner}
\end{figure}

\begin{figure}[htbp]
    \centering
    \includegraphics[width=1\linewidth]{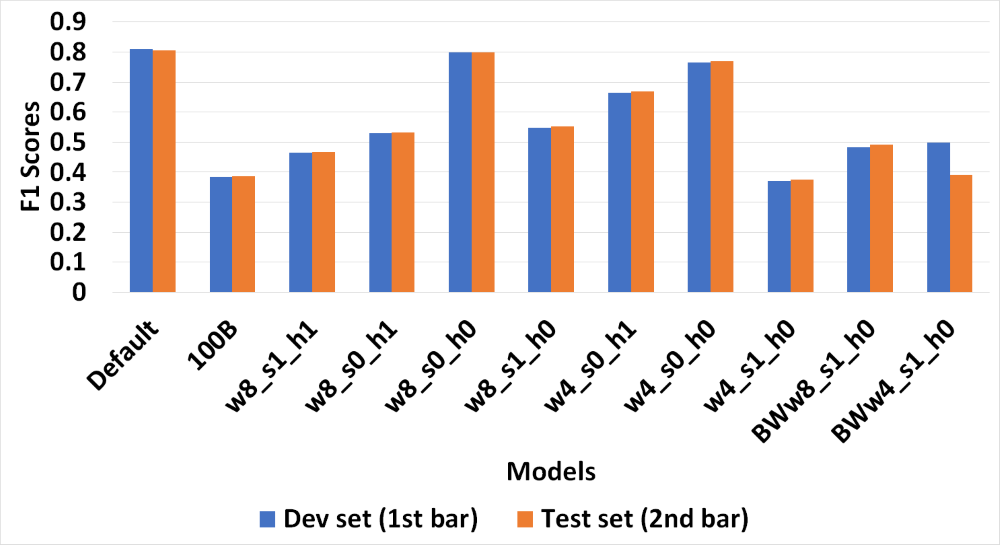}
    \caption{Sentiment Analysis (SA) Mean F1 Scores on IMDb Dataset}
    \label{fig:sa}
\end{figure}

The WordSim result output file from the Gensim Python program always has more than one value reported, including the Spearman correlation.
The first value is reported as WordSim score1 in the relevant table.
Table \ref{tab3} summarizes key results from the intrinsic evaluations for 300 dimensions\footnote{The results are to 3 decimal places}.
Table \ref{tab4} reveals the training time (in hours) and average embedding loading time (in seconds) representative of the various models used.
Tables \ref{tab5} and \ref{tab6} summarize key results for the extrinsic evaluations.
Figures \ref{fig:epoch5_wiki1},  \ref{fig:scores_corpora}, \ref{fig:corpora_300}, \ref{fig:ner} and \ref{fig:sa} present line graph of the eight combinations for different dimension sizes for SW, the trend of SW and BW corpora over several dimension sizes, analogy score comparison for models across datasets, NER mean F1 scores on the GMB dataset and SA mean F1 scores on the IMDb dataset, respectively.
Results for the smallest dataset (Wiki Abstract) are so poor, because of the tiny file size (15MB), there's no reason reporting them here.
Hence, we have focused on results from the SW and BW corpora.

Best combination in terms of analogy sometimes changes when corpus size increases, as will be noticed from table \ref{tab3}.
In terms of analogy score, for 10 epochs, w8s0h0 performs best while w8s1h0 performs best in terms of WordSim and corresponding Spearman correlation for SW.
Meanwhile, increasing the corpus size to BW, w4s1h0 performs best in terms of analogy score while w8s1h0 maintains its position as the best in terms of WordSim and Spearman correlation.
Besides considering quality metrics, it can be observed from table \ref{tab4} that comparative ratio of values between the models is not commensurate with the results in intrinsic or extrinsic values, especially when we consider the amount of time and energy spent, since more training time results in more energy consumption \cite{adewumi2019inner}.

\begin{table*}[!t]
\centering
\caption{Scores for 300 dimensions for 10 epochs for SW, BW \& 100B corpora.}
\begin{tabular}{c|c|c|c|c|c|c|c|c}
\textbf{} & \footnotesize{\textbf{w8s1h1}} & \footnotesize{\textbf{w8s0h1}} & \footnotesize{\textbf{w8s0h0}} &
\footnotesize{\textbf{w8s1h0}} & \footnotesize{\textbf{w4s1h1}} & \footnotesize{\textbf{w4s0h1}} & \footnotesize{\textbf{w4s0h0}} & \footnotesize{\textbf{w4s1h0}} \\
\hline
\multicolumn{9}{|c|}{\footnotesize{Simple Wiki}} \\
\hline
\footnotesize{Analogy} & \footnotesize{0.461} & \footnotesize{0.269} & \footnotesize{\textbf{0.502}} & \footnotesize{0.439} & \footnotesize{0.446} & \footnotesize{0.243} & \footnotesize{0.478} & \footnotesize{0.407} \\
\hline
\footnotesize{WordSim score1} & \footnotesize{0.636} & \footnotesize{0.611} & \footnotesize{0.654} & \footnotesize{\textbf{0.655}} & \footnotesize{0.635} & \footnotesize{0.608} & \footnotesize{0.620} & \footnotesize{0.635} \\
\hline
\footnotesize{Spearman} & \footnotesize{0.670} & \footnotesize{0.648} & \footnotesize{0.667} & \footnotesize{\textbf{0.695}} & \footnotesize{0.668} & \footnotesize{0.648} & \footnotesize{0.629} & \footnotesize{0.682} \\
\hline
\multicolumn{9}{|c|}{\footnotesize{Billion Word}} \\
\hline
\footnotesize{Analogy} &
\footnotesize{0.587} & \footnotesize{0.376} &
\footnotesize{0.638} & \footnotesize{0.681} & \footnotesize{0.556} & \footnotesize{0.363} & \footnotesize{0.629} & \footnotesize{\textbf{0.684}} \\
\hline
\footnotesize{WordSim score1} &
\footnotesize{0.614} & \footnotesize{0.511} &
\footnotesize{0.599} & \footnotesize{\textbf{0.644}} & \footnotesize{0.593} & \footnotesize{0.508} & \footnotesize{0.597} & \footnotesize{0.635} \\
\hline
\footnotesize{Spearman} &
\footnotesize{0.653} & \footnotesize{0.535} & \footnotesize{0.618} & \footnotesize{\textbf{0.681}} & \footnotesize{0.629} & \footnotesize{0.527} & \footnotesize{0.615} & \footnotesize{0.677} \\
\hline
\multicolumn{9}{|c|}{\footnotesize{Google News - 100B (s1h0)}} \\
\hline
\multicolumn{3}{c}{\footnotesize{Analogy: 0.740}} &
\multicolumn{3}{c}{\footnotesize{WordSim: 0.624}} & \multicolumn{3}{c}{\footnotesize{Spearman: 0.659}} \\
\hline
\end{tabular}
\label{tab3}
\end{table*}

\begin{table}[h]
\centering
\caption{ Training \& embedding loading time for w8s1h0, w8s1h1 \& 100B}
\begin{tabular}{c|c|c}
\footnotesize{\textbf{Model}} & \footnotesize{\textbf{Training (hours)}} &  \footnotesize{\textbf{Loading Time (s)}}\\
\hline
\footnotesize{SW w8s1h0} & \footnotesize{5.44} &  \footnotesize{1.93}\\
\hline
\footnotesize{BW w8s1h1} & \footnotesize{27.22} &  \footnotesize{4.89}\\
\hline
\footnotesize{GoogleNews (100B)} & \footnotesize{-} &  \footnotesize{97.73}\\
\hline
\end{tabular}
\label{tab4}
\end{table}

Information on the length of training time for the original 100B model is not readily available.
However, it's interesting to note that it is a skipgram-negative sampling (s1h0) model.
Its analogy score, which we tested and report, is confirmed in the original paper \cite{mikolov2013efficient}.
It beats our best models in only analogy score (even for SW), performing worse in others, despite using a much bigger corpus of 3,000,000 vocabulary size and 100 billion words while SW had vocabulary size of 367,811 and is 711MB.
It is very likely our analogy scores will improve when we use a much larger corpus, as can be observed from table \ref{tab3}, which involves just one billion words.

With regards to increasing dimension, though the two best combinations in analogy (w8s0h0 \& w4s0h0) for SW, as shown in fig. \ref{fig:epoch5_wiki1}, decreased only slightly compared to others, the increased training time and much larger serialized model size render any possible minimal score advantage with higher dimensions undesirable.
As can be observed in fig. \ref{fig:scores_corpora}, from 100 dimensions, scores improve but start to drop after over 300 dimensions for SW and after over 400 dimensions for BW, confirming the observation by \cite{mikolov2013efficient}.
This trend is true for all combinations for all tests.
Polynomial interpolation may be used to determine the optimal dimension in both corpora.

\begin{table}[htbp!]
\centering
\caption{NER Dev and Test sets Mean Results}
\resizebox{\columnwidth}{!}{%
\begin{tabular}{c|c|c|c|c|c}
\footnotesize{\textbf{Metric}} &
\footnotesize{\textbf{Default}} & \footnotesize{\textbf{100B}} & \footnotesize{\textbf{w8 s0 h0}} & \footnotesize{\textbf{w8 s1 h0}} &
\footnotesize{\textbf{BW w4 s1 h0}}\\
\hline
\footnotesize{} &
\footnotesize{Dev, Test} & \footnotesize{Dev, Test} & \footnotesize{Dev, Test} & \footnotesize{Dev, Test} &
\footnotesize{Dev, Test}\\
\hline
\footnotesize{F1} &
\footnotesize{0.661, 0.661} & \footnotesize{\textbf{0.679}, 0.676} & \footnotesize{0.668, 0.669} & \footnotesize{0.583, 0.676} &
\footnotesize{\textbf{0.679, 0.677}}\\
\hline
\footnotesize{Precision} &
\footnotesize{0.609, 0.608} & \footnotesize{\textbf{0.646, 0.642}} & \footnotesize{0.636, 0.637} & \footnotesize{0.553, 0.642} &
\footnotesize{0.644, \textbf{0.642}}\\
\hline
\footnotesize{Recall} &
\footnotesize{0.723, \textbf{0.724}} & \footnotesize{0.716, 0.714} & \footnotesize{0.704, 0.706} & \footnotesize{0.618, 0.715} &
\footnotesize{0.717, 0.717}\\
\hline
\end{tabular}
}
\label{tab5}
\end{table}

\begin{table}[htbp!]
\centering
\caption{Sentiment Analysis Dev and Test sets Mean Results}
\resizebox{\columnwidth}{!}{%
\begin{tabular}{c|c|c|c|c|c}
\textbf{Metric} &
\textbf{Default} & \textbf{100B} & \textbf{w8 s0 h0} & \textbf{w8 s1 h0} &
\textbf{BW w4 s1 h0}\\
\hline
 &
Dev, Test & Dev, Test & Dev, Test & Dev, Test &
Dev, Test\\
\hline
F1 &
\textbf{0.810, 0.805} & 0.384, 0.386 & 0.798, 0.799 & 0.548, 0.553 &
0.498, 0.390\\
\hline
Precision &
0.805, 0.795 & 0.6, 0.603 & \textbf{0.814, 0.811} & 0.510, 0.524 &
0.535, 0.533\\
\hline
Recall &
\textbf{0.818, 0.816} & 0.303, 0.303 & 0.788, 0.792 & 0.717, 0.723 &
0.592, 0.386\\
\hline
Accuracy &
\textbf{0.807, 0.804} & 0.549, 0.55 & 0.801, 0.802 & 0.519, 0.522 &
0.519, 0.517\\
\hline
\end{tabular}
}
\label{tab6}
\end{table}

With regards to NER, most pretrained embeddings outperformed the default Pytorch embedding, with our BW w4s1h0 model (which is best in BW analogy score) performing best in F1 score and closely followed by the 100B model.
On the other hand, with regards to SA, Pytorch embedding outperformed the pretrained embeddings but was closely followed by our SW w8s0h0 model (which also had the best SW analogy score).
100B performed second worst of all, despite originating from a very huge corpus.
The combinations w8s0h0 \& w4s0h0 of SW performed reasonably well in both extrinsic tasks, just as the default Pytorch embedding did.

Significance tests using bootstrap, based on \cite{calmettes2012making}, on the results of the differences in means of the 100B \& BW w4s1h0 models for NER shows a 95\% confidence interval (CI) of [-0.008, 0.01] but [0.274, 0.504] for 100B \& SW w8s0h0 for SA.
Since one algorithm is involved in the comparisons in each case, unlike multiple algorithms \cite{demvsar2006statistical}, the applied bootstrap approach is adequate.
The CI interval for NER includes 0, thus we can conclude the difference was likely due to chance and fail to reject the null hypothesis but the CI for SA does not include 0, thus the difference is unlikely due to chance so we reject the null hypothesis.

\section{Conclusions}
This work analyses, empirically,  optimal combinations of hyper-parameters for embeddings, specifically for word2vec.
It further shows that for downstream tasks, like NER and SA, there's no silver bullet!
However, some combinations show strong performance across tasks.
Performance of embeddings is task-specific and high analogy scores do not necessarily correlate positively with performance on downstream tasks.
This point on correlation is somewhat similar to results by \cite{chiu2016intrinsic} and \cite{wang2019evaluating}.
It was discovered that increasing embedding dimension size depreciates performance after a point.
If strong considerations of saving time, energy and the environment are made, then reasonably smaller corpora may suffice or even be better in some cases.
The on-going drive by many researchers to use ever-growing data to train deep neural networks can benefit from the findings of this work.
Indeed, hyper-parameter choices are very important in neural network systems \cite{levy2015improving}.

Future work that may be investigated are the performance of other architectures of word or sub-word embeddings in SotA networks like BERT (based on a matrix of hyper-parameters), the performance and comparison of embeddings applied to other less-explored languages, and how these embeddings perform in other downstream tasks.

\section*{Funding}
This work was supported partially by Vinnova under the project number 2019-02996 'Språkmodeller för svenska myndigheter'. They, however, had no involvement in any stage of this work, including study design, interpretation of data and report writing.



\bibliographystyle{unsrt}
\bibliography{ref}

\end{document}